\documentclass[11pt]{article}
\usepackage{acl2016}
\usepackage{times}
\usepackage{latexsym}

\usepackage{amsmath, amssymb}
\usepackage{bm}
\usepackage[pdftex]{graphicx}
\usepackage{url}
\usepackage{multirow}

\aclfinalcopy 

\title{Adaptive Joint Learning of \\ Compositional and Non-Compositional Phrase Embeddings}

\author{
	Kazuma Hashimoto and Yoshimasa Tsuruoka\\
	The University of Tokyo, 3-7-1 Hongo, Bunkyo-ku, Tokyo, Japan\\
	{\tt \{hassy,tsuruoka\}@logos.t.u-tokyo.ac.jp}\\
}

\date{}

\begin{document}

\maketitle

\begin{abstract}
We present a novel method for jointly learning compositional and non-compositional phrase embeddings by adaptively weighting both types of embeddings using a compositionality scoring function.
The scoring function is used to quantify the level of compositionality of each phrase, and the parameters of the function are jointly optimized with the objective for learning phrase embeddings.
In experiments, we apply the adaptive joint learning method to the task of learning embeddings of transitive verb phrases, and show that the compositionality scores have strong correlation with human ratings for verb-object compositionality, substantially outperforming the previous state of the art.
Moreover, our embeddings improve upon the previous best model on a transitive verb disambiguation task.
We also show that a simple ensemble technique further improves the results for both tasks.
\end{abstract}

\section{Introduction}

Representing words and phrases in a vector space has proven effective in a variety of language processing tasks~\cite{cphrase,seq2seq}.
In most of the previous work, phrase embeddings are computed from word embeddings by using various kinds of composition functions.
Such composed embeddings are called {\it compositional embeddings}.
An alternative way of computing phrase embeddings is to treat phrases as single units and assigning a unique embedding to each  candidate phrase~\cite{word2vec,yazdani2015}.
Such embeddings are called {\it non-compositional embeddings}.

Relying solely on non-compositional embeddings has the obvious problem of data sparsity (i.e. rare or unknown phrase problems).
At the same time, however, using compositional embeddings is not always the best option since some phrases are inherently non-compositional.
For example, the phrase ``bear fruits'' means ``to yield results''\footnote{The definition is found at \url{http://idioms.thefreedictionary.com/bear+fruit}.} but it is hard to infer its meaning by composing the meanings of ``bear'' and ``fruit''.
Treating all phrases as compositional also has a negative effect in learning the composition function because the words in those idiomatic phrases are not just uninformative but can serve as noisy samples in the training.
These problems have motivated us to adaptively combine both types of embeddings.

Most of the existing methods for learning phrase embeddings can be divided into two approaches.
One approach is to learn compositional embeddings by regarding all phrases as compositional~\cite{cphrase,socher2012}.
The other approach is to learn both types of embeddings separately and use the better ones~\cite{kartsaklis2014,muraoka2014}.
\newcite{kartsaklis2014} show that non-compositional embeddings are better suited for a phrase similarity task,
whereas \newcite{muraoka2014} report the opposite results on other tasks.
These results suggest that we should not stick to either of the two types of embeddings unconditionally and could learn better phrase embeddings by considering the compositionality levels of the individual phrases in a more flexible fashion.

In this paper, we propose a method that jointly learns compositional and non-compositional embeddings by adaptively weighting both types of phrase embeddings using a compositionality scoring function.
The scoring function is used to quantify the level of compositionality of each phrase and learned in conjunction with the target task for learning phrase embeddings.
In experiments, we apply our method to the task of learning transitive verb phrase embeddings and demonstrate that it allows us to achieve state-of-the-art performance on standard datasets for compositionality detection and verb disambiguation.

\section{Method}

\begin{figure}[t]
  \begin{center}
    \includegraphics[width=80mm]{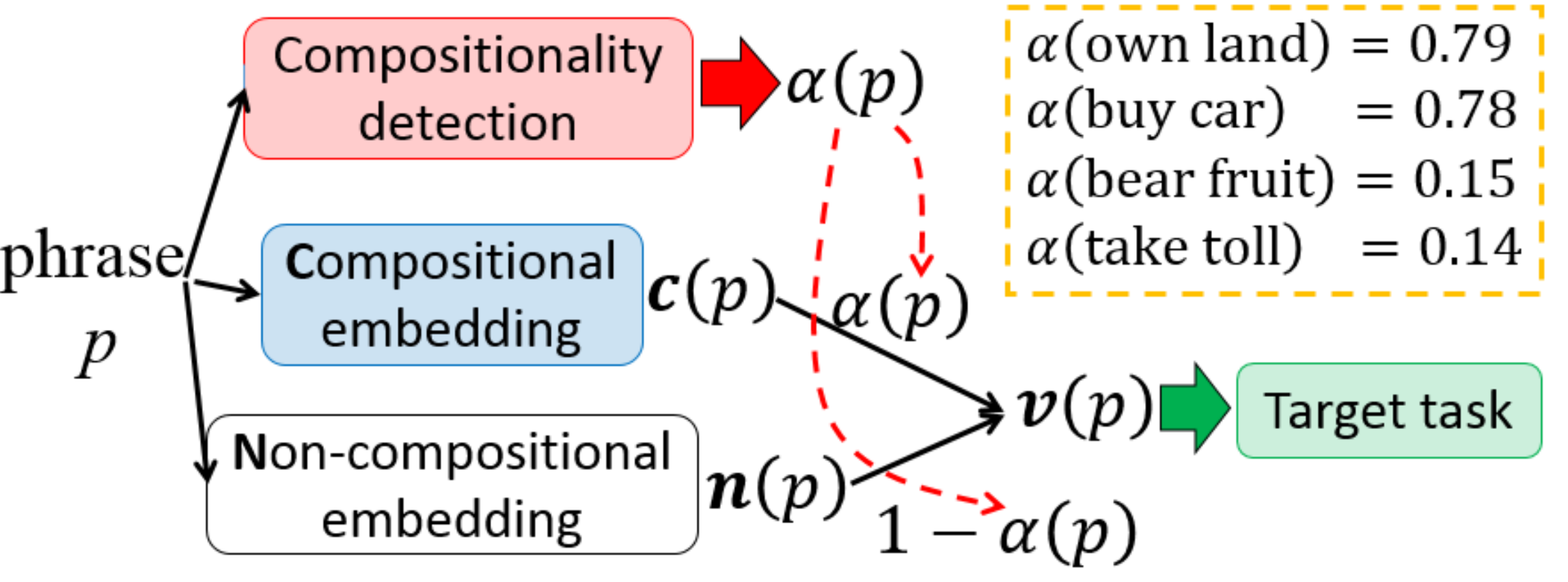}
    \caption{The overview of our method and examples of the compositionality scores.
			Given a phrase $p$, our method first computes the compositionality score $\alpha(p)$ (Eq.~(\ref{eq:feature})), and then computes the phrase embedding $\bm{v}(p)$ using the compositional and non-compositional embeddings, $\bm{c}(p)$ and $\bm{n}(p)$, respectively (Eq.~(\ref{eq:balance})).
			}
    \label{fig:overview}
  \end{center}
\end{figure}

In this section, we describe our approach in the most general form, without specifying the function to compute the compositional embeddings or the target task for optimizing the embeddings.

Figure~\ref{fig:overview} shows the overview of our proposed method.
At each iteration of the training (i.e. gradient calculation) of a certain target task (e.g. language modeling or sentiment analysis), our method first computes a compositionality score for each phrase.
Then the score is used to weight the compositional and non-compositional embeddings of the phrase in order to compute the expected embedding of the phrase which is to be used in the target task.
Some examples of the compositionality scores are also shown in the figure.

\subsection{Compositional Phrase Embeddings}
The compositional embedding $\bm{c}(p)\in\mathbb{R}^{d\times 1}$ of a phrase $p=(w_1, \cdots, w_L)$ is formulated as
\begin{equation}
\bm{c}(p)=f(\bm{v}(w_1), \cdots, \bm{v}(w_L)) ,
\end{equation}
where $d$ is the dimensionality, $L$ is the phrase length, $\bm{v}(\cdot)\in\mathbb{R}^{d\times 1}$ is a word embedding, and $f(\cdot)$ is a composition function.
The function can be simple ones such as element-wise addition or multiplication~\cite{lapata2008}.
More complex ones such as recurrent neural networks~\cite{seq2seq} are also commonly used.
The word embeddings and the composition function are jointly learned on a certain target task.
Since compositional embeddings are built on word-level (i.e. unigram) information, they are less prone to the data sparseness problem.

\subsection{Non-Compositional Phrase Embeddings}
In contrast to the compositional embedding, the non-compositional embedding of a phrase $\bm{n}(p)\in\mathbb{R}^{d\times 1}$ is independently parameterized, i.e., the phrase $p$ is treated just like a single word.
\newcite{word2vec} show that non-compositional embeddings are preferable when dealing with idiomatic phrases.
Some recent studies~\cite{kartsaklis2014,muraoka2014} have discussed the (dis)advantages of using compositional or non-compositional embeddings.
However, in most cases, a phrase is neither completely compositional nor completely non-compositional.
To the best of our knowledge, there is no method that allows us to jointly learn both types of phrase embeddings by incorporating the levels of compositionality of the phrases as real-valued scores.

\subsection{Adaptive Joint Learning}
To simultaneously consider both compositional and non-compositional aspects of each phrase, we compute a phrase embedding $\bm{v}(p)$ by adaptively weighting $\bm{c}(p)$ and $\bm{n}(p)$ as follows:
\begin{equation}
\label{eq:balance}
\bm{v}(p)=\alpha(p)\bm{c}(p)+(1-\alpha(p))\bm{n}(p) ,
\end{equation}
where $\alpha(\cdot)$ is a scoring function that quantifies the compositionality levels,
and outputs a real value ranging from 0 to 1.
What we expect from the scoring function is that large scores indicate high levels of compositionality.
In other words, when $\alpha(p)$ is close to 1, the compositional embedding is mainly considered, and vice versa.
For example, we expect $\alpha(\mathrm{buy~car})$ to be large and $\alpha(\mathrm{bear~fruit})$ to be small as shown in Figure~\ref{fig:overview}.

We parameterize the scoring function $\alpha(p)$ as logistic regression:
\begin{equation}
\label{eq:feature}
\alpha(p)=\sigma(\bm{W}\cdot\boldsymbol{\phi}(p)) ,
\end{equation}
where $\boldsymbol{\phi}(p)\in\mathbb{R}^{N\times 1}$ is a feature vector of the phrase $p$, $\bm{W}\in\mathbb{R}^{N\times1}$ is a weight vector, $N$ is the number of features, and $\sigma(\cdot)$ is the logistic function.
The weight vector $\bm{W}$ is jointly optimized in conjunction with the objective $J$ for the target task of learning phrase embeddings $\bm{v}(p)$.

\paragraph{Updating the model parameters}
Given the partial derivative $\boldsymbol{\delta}_{p}=\frac{\partial{J}}{\partial{\bm{v}(p)}}\in\mathbb{R}^{d\times 1}$ for the target task, we can compute the partial derivative for updating $\bm{W}$ as follows:
\begin{align}
&\delta_\alpha=\alpha(p)(1-\alpha(p))\{\boldsymbol{\delta}_{p}\cdot(\bm{c}(p)-\bm{n}(p))\}\\
&\frac{\partial{J}}{\partial{\bm{W}}}=\delta_\alpha\boldsymbol{\phi}(p) .
\end{align}
If $\boldsymbol{\phi}(p)$ is not constructed by static features but is computed by a feature learning model such as neural networks, we can propagate the error term $\delta_\alpha$ into the feature learning model by the following equation:
\begin{equation}
\frac{\partial{J}}{\partial{\boldsymbol{\phi}(p)}}=\delta_\alpha\bm{W} .
\end{equation}
When we use only static features, as in this work, we can simply compute the partial derivatives of $J$ with respect to $\bm{c}(p)$ and $\bm{n}(p)$ as follows:
\begin{align}
\frac{\partial{J}}{\partial{\bm{c}(p)}}&=\alpha(p)\boldsymbol{\delta}_p \label{eq:update_c}\\
\frac{\partial{J}}{\partial{\bm{n}(p)}}&=(1-\alpha(p))\boldsymbol{\delta}_p \label{eq:update_n} .
\end{align}
As mentioned above, Eq.~(\ref{eq:update_c}) and (\ref{eq:update_n}) show that the non-compositional embeddings are mainly updated when $\alpha(p)$ is close to 0, and vice versa.
The partial derivative $\frac{\partial{J}}{\partial{\bm{c}(p)}}$ is used to update the model parameters in the composition function via the backpropagation algorithm.
Any differentiable composition functions can be used in our method.

\paragraph{Expected behavior of our method}
The training of our method depends on the target task;
that is, the model parameters are updated so as to minimize the cost function as described above.
More concretely, $\alpha(p)$ for each phrase $p$ is adaptively adjusted so that the corresponding parameter updates contribute to minimizing the cost function.
As a result, different phrases will have different $\alpha(p)$ values depending on their compositionality.
If the size of the training data were almost infinitely large, $\alpha(p)$ for all phrases would become nearly zero, and the non-compositional embeddings $\bm{n}(p)$ are dominantly used (since that would allow the model to better fit the data).
In reality, however, the amount of the training data is limited, and thus the compositional embeddings $\bm{c}(p)$ are effectively used to overcome the data sparseness problem.

\section{Learning Verb Phrase Embeddings}
This section describes a particular instantiation of our approach presented in the previous section, focusing on the task of learning the embeddings of transitive verb phrases.

\subsection{Word and Phrase Prediction in Predicate-Argument Relations}
\label{subsec:task}
Acquisition of selectional preference using embeddings has been widely studied, where word and/or phrase embeddings are learned based on syntactic links~\cite{bansal2014,hashimoto2015,levy2014,vande2014}.
As with language modeling, these methods perform word (or phrase) prediction using (syntactic) contexts.

In this work, we focus on verb-object relationships and employ a phrase embedding learning method presented in \newcite{hashimoto2015}.
The task is a plausibility judgment task for predicate-argument tuples.
They extracted Subject-Verb-Object ({\it SVO}) and SVO-Preposition-Noun ({\it SVOPN}) tuples using a probabilistic HPSG parser, {\it Enju}~\cite{miyao2008}, from the training corpora.
Transitive verbs and prepositions are extracted as predicates with two arguments.
For example, the extracted tuples include (S, V, O) = (``importer'', ``make'', ``payment'') and (SVO, P, N) = (``importer make payment'', ``in'', ``currency'').
The task is to discriminate between observed and unobserved tuples, such as the (S, V, O) tuple mentioned above and (S, V', O) = (``importer'', ``{\it eat}'', ``payment''), which is generated by replacing ``make'' with ``eat''.
The (S, V', O) tuple is unlikely to be observed.

For each tuple $(p, a_1, a_2)$ observed in the training data, a cost function is defined as follows:
\begin{equation}
\label{eq:cost_func}
\begin{split}
-\log{\sigma(s(p, a_1, a_2))}-&\log{\sigma(-s(p', a_1, a_2))}\\
-&\log{\sigma(-s(p, a_1', a_2))}\\
-&\log{\sigma(-s(p, a_1, a_2'))} ,
\end{split}
\end{equation}
where $s(\cdot)$ is a plausibility scoring function, and $p$, $a_1$ and $a_2$ are a predicate and its arguments, respectively.
Each of the three unobserved tuples $(p', a_1, a_2)$, $(p, a_1', a_2)$, and $(p, a_1, a_2')$ is generated by replacing one of the entries with a random sample.

In their method, each predicate $p$ is represented with a matrix $\bm{M}(p)\in\mathbb{R}^{d\times d}$ and each argument $a$ with an embedding $\bm{v}(a)\in\mathbb{R}^{d\times 1}$.
The matrices and embeddings are learned by minimizing the cost function using {\it AdaGrad}~\cite{duchi2011}.
The scoring function is parameterized as
\begin{equation}
s(p, a_1, a_2)=\bm{v}(a_1)\cdot(\bm{M}(p)\bm{v}(a_2)) ,
\end{equation}
and the VO and SVO embeddings are computed as
\begin{align}
\bm{v}(VO)&=\bm{M}(V)\bm{v}(O) \label{eq:copy_obj} \\
\bm{v}(SVO)&=\bm{v}(S)\odot\bm{v}(VO) \label{eq:svo_embed} ,
\end{align}
as proposed by \newcite{kart2012}.
The operator $\odot$ denotes element-wise multiplication.
In summary, the scores are computed as
\begin{align}
s(V, S, O)&=\bm{v}(S)\cdot\bm{v}(VO) \label{eq:score_svo} \\
s(P, SVO, N)&=\bm{v}(SVO)\cdot(\bm{M}(P)\bm{v}(N)) \label{eq:score_svopn} .
\end{align}
With this method, the word and composed phrase embeddings are jointly learned based on co-occurrence statistics of predicate-argument structures.
Using the learned embeddings, they achieved state-of-the-art accuracy on a transitive verb disambiguation task~\cite{grefenstette2011}.

\subsection{Applying the Adaptive Joint Learning}
In this section, we apply our adaptive joint learning method to the task described in Section~\ref{subsec:task}.
We here redefine the computation of $\bm{v}(VO)$ by first replacing $\bm{v}(VO)$ in Eq.~(\ref{eq:copy_obj}) with $\bm{c}(VO)$ as,
\begin{equation}
\bm{c}(VO)=\bm{M}(V)\bm{v}(O) ,
\end{equation}
and then assigning $VO$ to $p$ in Eq.~(\ref{eq:balance}) and (\ref{eq:feature}):
\begin{align}
\bm{v}(VO)&=\alpha(VO)\bm{c}(VO)+(1-\alpha(VO))\bm{n}(VO) \label{eq:balance_VO} ,\\
\alpha(VO)&=\sigma(\bm{W}\cdot\boldsymbol{\phi}(VO)) .
\end{align}
The $\bm{v}(VO)$ in Eq.~(\ref{eq:balance_VO}) is used in Eq.~(\ref{eq:svo_embed}) and (\ref{eq:score_svo}).
We assume that the candidates of the phrases are given in advance.
For the phrases not included in the candidates, we set $\bm{v}(VO)=\bm{c}(VO)$.
This is analogous to the way a human guesses the meaning of an idiomatic phrase she does not know.
We should note that $\boldsymbol{\phi}(VO)$ can be computed for phrases not included in the candidates, using partial features among the features described below.
If any features do not fire, $\boldsymbol{\phi}(VO)$ becomes 0.5 according to the logistic function.

For the feature vector $\boldsymbol{\phi}(VO)$, we use the following simple binary and real-valued features:
\begin{itemize}
\item indices of V, O, and VO
\item frequency and Pointwise Mutual Information (PMI) values of VO.
\end{itemize}
More concretely, the first set of the features (indices of V, O, and VO) is the concatenation of traditional one-hot vectors.
The second set of features, frequency and PMI~\cite{pmi} features, have proven effective in detecting the compositionality of transitive verbs in \newcite{mccarthy2007} and \newcite{venkatapathy2005}.
Given the training corpus, the frequency feature for a VO pair is computed as
\begin{equation}
freq(VO)=\log(count(VO)) ,
\end{equation}
where $count(VO)$ counts how many times the VO pair appears in the training corpus,
and the PMI feature is computed as
\begin{equation}
\mathrm{PMI}(VO)= \log{\frac{count(VO)count(*)}{count(V)count(O)}},
\end{equation}
where $count(V)$, $count(O)$, and $count(*)$ are the counts of the verb $V$, the object $O$, and all VO pairs in the training corpus, respectively.
We normalize the frequency and PMI features so that their maximum absolute value becomes 1.

\section{Experimental Settings}

\subsection{Training Data}
As the training data, we used two datasets, one small and one large: the British National Corpus (BNC)~\cite{bnc} and the English Wikipedia.
More concretely, we used the publicly available data\footnote{\url{http://www.logos.t.u-tokyo.ac.jp/~hassy/publications/cvsc2015/}} preprocessed by \newcite{hashimoto2015}.
The BNC data consists of 1.38 million SVO tuples and 0.93 million SVOPN tuples.
The Wikipedia data consists of 23.6 million SVO tuples and 17.3 million SVOPN tuples.
Following the provided code\footnote{\url{https://github.com/hassyGo/SVOembedding}}, we used exactly the same train/development/test split (0.8/0.1/0.1) for training the overall model.
As the third training data, we also used the concatenation of the two data, which is hereafter referred to as {\it BNC-Wikipedia}.

We applied our adaptive joint learning method to verb-object phrases observed more than $K$ times in each corpus.
$K$ was set to 10 for the BNC data and 100 for the Wikipedia and BNC-Wikipedia data.
Consequently, the non-compositional embeddings were assigned to 17,817, 28,933, and 30,682 verb-object phrase types in the BNC, Wikipedia, and BNC-Wikipedia data, respectively.

\subsection{Training Details}
The model parameters consist of $d$-dimensional word embeddings for nouns, non-compositional phrase embeddings, 
$d$$\times$$d$-dimensional matrices for verbs and prepositions, and a weight vector $\bm{W}$ for $\alpha(VO)$.
All the model parameters are jointly optimized.
We initialized the embeddings and matrices with zero-mean gaussian random values with a variance of $\frac{1}{d}$ and $\frac{1}{d^2}$, respectively, and $\bm{W}$ with zeros.
Initializing $\bm{W}$ with zeros forces the initial value of each $\alpha(VO)$ to be $0.5$ since we use the logistic function to compute $\alpha(VO)$.

The optimization was performed via mini-batch AdaGrad~\cite{duchi2011}.
We fixed $d$ to $25$ and the mini-batch size to $100$.
We set candidate values for the learning rate $\varepsilon$ to $\{0.01, 0.02, 0.03, 0.04, 0.05\}$.
For the weight vector $\bm{W}$, we employed L2-norm regularization and set the coefficient $\lambda$ to $\{10^{-3}, 10^{-4}, 10^{-5}, 10^{-6}, 0\}$.
For selecting the hyperparameters, each training process was stopped when the evaluation score on the development split decreased.
Then the best performing hyperparameters were selected for each training dataset.
Consequently, $\varepsilon$ was set to $0.05$ for all training datasets, and $\lambda$ was set to $10^{-6}$, $10^{-3}$, and $10^{-5}$ for the BNC, Wikipedia, and BNC-Wikipedia data, respectively.
Once the training is finished, we can use the learned embeddings and the scoring function in downstream target tasks.

\section{Evaluation on the Compositionality Detection Function}

\subsection{Evaluation Settings}
\label{subsec:compos_setting}

\paragraph{Datasets}
First, we evaluated the learned compositionality detection function on two datasets, {\bf VJ'05}\footnote{\url{http://www.dianamccarthy.co.uk/downloads/SVAJ2005compositionality_rating.txt}} and {\bf MC'07}\footnote{\url{http://www.dianamccarthy.co.uk/downloads/emnlp2007data.txt}}, provided by \newcite{venkatapathy2005} and \newcite{mccarthy2007}, respectively.
VJ'05 consists of 765 verb-object pairs with human ratings for the compositionality.
MC'07 is a subset of VJ'05 and consists of 638 verb-object pairs.
For example, the rating of ``buy car'' is 6, which is the highest score, indicating the phrase is highly compositional.
The rating of ``bear fruit	'' is 1, which is the lowest score, indicating the phrase is highly non-compositional.

\paragraph{Evaluation metric}
The evaluation was performed by calculating Spearman's rank correlation scores\footnote{We used the Scipy 0.12.0 implementation in Python.} between the averaged human ratings and the learned compositionality scores $\alpha(VO)$.

\paragraph{Ensemble technique}
We also produced the result by employing an {\it ensemble} technique.
More concretely, we used the averaged compositionality scores from the results of the BNC and Wikipedia data for the ensemble result.

\subsection{Results and Discussion}

\subsubsection{Result Overview}
\label{subsubsec:detect_res}

\begin{table}[t]
  \begin{center}
{\small
    \begin{tabular}{l|c|c}
      Method & MC'07 & VJ'05  \\ \hline \hline

     Proposed method (Wikipedia) & 0.508 & 0.514 \\
     Proposed method (BNC) & 0.507 & 0.507 \\
     Proposed method (BNC-Wikipedia) & 0.518 & 0.527 \\
	Proposed method (Ensemble) & 0.550 & 0.552 \\ \hline

	\newcite{kiela2013} w/ WordNet & n/a & 0.461 \\
	\newcite{kiela2013} & n/a & 0.420 \\
	DSPROTO~\cite{mccarthy2007} & 0.398 & n/a \\
	PMI~\cite{mccarthy2007} & 0.274 & n/a \\
 	Frequency~\cite{mccarthy2007} & 0.141 & n/a \\ \hline

	DSPROTO+~\cite{mccarthy2007} & 0.454 & n/a \\ \hline
	Human agreement & 0.702 & 0.716 \\ \hline
    \end{tabular}
}
    \caption{Compositionality detection task.}
    \label{tb:compos}
  \end{center}
\end{table}

Table~\ref{tb:compos} shows our results and the state of the art.
Our method outperforms the previous state of the art in all settings.
The result denoted as {\it Ensemble} is the one that employs the ensemble technique, and achieves the strongest correlation with the human-annotated datasets.
Even without the ensemble technique, our method performs better than all of the previous methods.

\newcite{kiela2013} used window-based co-occurrence vectors and improved their score using WordNet hypernyms.
By contrast, our method does not rely on such external resources, and only needs parsed corpora.
We should note that \newcite{kiela2013} reported that their score did not improve when using parsed corpora.
Our method also outperforms DSPROTO+, which used a small amount of the labeled data, while our method is fully unsupervised.

We calculated confidence intervals ($P<0.05$) using bootstrap resampling~\cite{bootstrap}.
For example, for the results using the BNC-Wikipedia data, the intervals on MC'07 and VJ'05 are (0.455, 0.574) and (0.475, 0.579), respectively.
These results show that our method significantly outperforms the previous state-of-the-art results.

\begin{table*}[t]
  \begin{center}
{\small
    \begin{tabular}{ll|c|c|c|c|c}
     \multicolumn{2}{l|}{Phrase} & Gold standard & (a) BNC & (b) Wikipedia & BNC-Wikipedia & Ensemble ((a)+(b))$\times$0.5 \\ \hline \hline
	\multirow{5}{*}{(A)}
	& buy car & 6 & 0.78 & 0.71 & 0.80 & 0.74 \\
	& own land & 6 & 0.79 & 0.73 & 0.76 & 0.76 \\
	& take toll & 1.5 & 0.14 & 0.11 & 0.06 & 0.13 \\
	& shed light & 1 & 0.21 & 0.07 & 0.07 & 0.14 \\
	& bear fruit & 1 & 0.15 & 0.19 & 0.17 & 0.17 \\ \hline
	\multirow{2}{*}{(B)}
	&make noise & 6 & 0.37 & 0.33 & 0.30 & 0.35 \\
	& have reason & 5 & 0.26 & 0.39 & 0.33 & 0.33 \\ \hline
	\multirow{2}{*}{(C)}
	& smoke cigarette & 6 & 0.56 & 0.90 & 0.78 & 0.73 \\
	& catch eye & 1 & 0.48 & 0.14 & 0.17 & 0.31 \\ \hline
    \end{tabular}
}
    \caption{Examples of the compositionality scores.}
    \label{tb:examples}
  \end{center}
\end{table*}

\subsubsection{Analysis of Compositionality Scores}

\begin{figure}[t]
  \begin{center}
    \includegraphics[width=82mm]{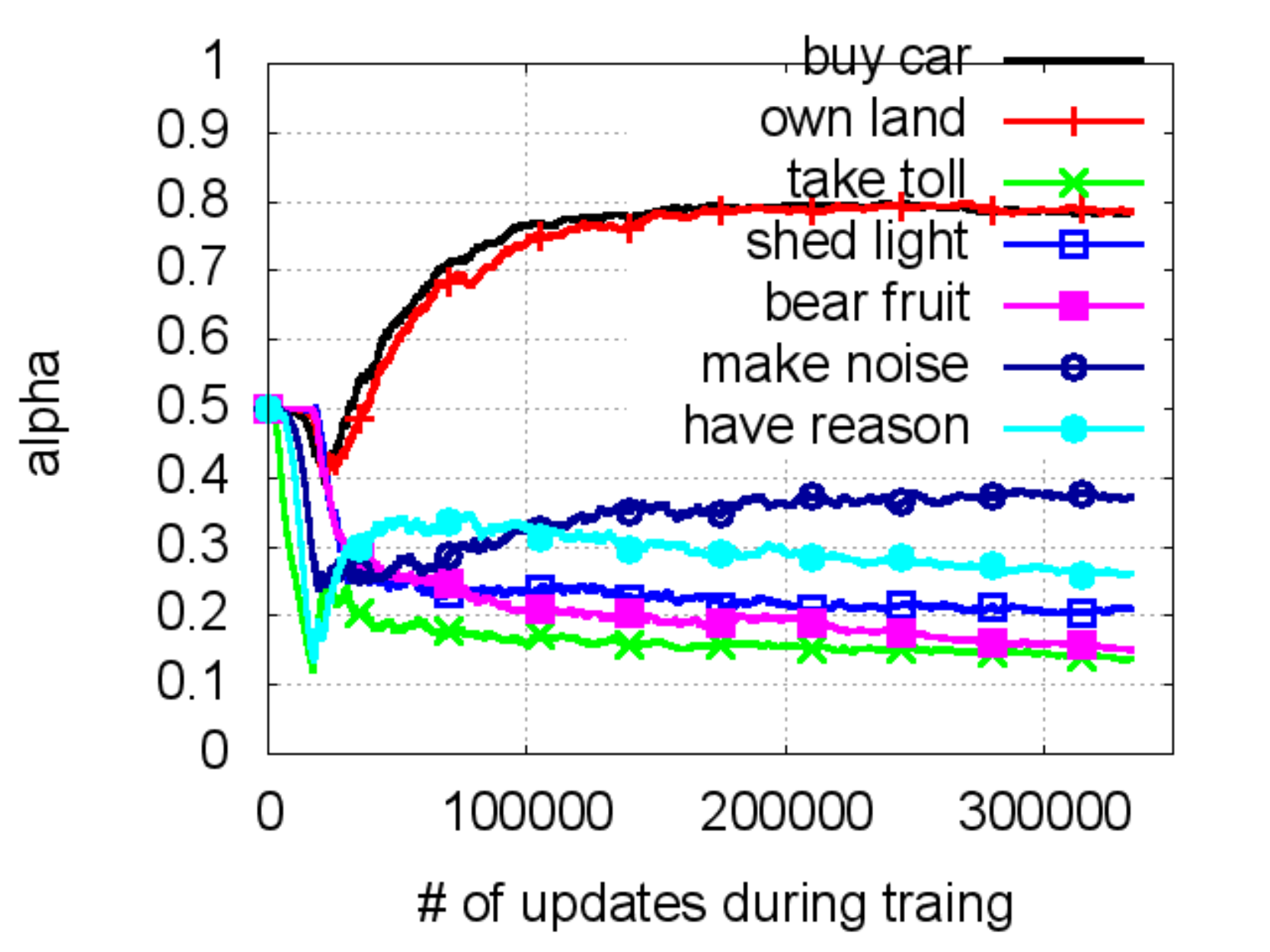}
    \caption{Trends of $\alpha(VO)$ during the training on the BNC data.}
    \label{fig:alpha}
  \end{center}
\end{figure}

Figure~\ref{fig:alpha} shows how $\alpha(VO)$ changes for the seven phrases during the training on the BNC data.
As shown in the figure, starting from $0.5$, $\alpha(VO)$ for each phrase converges to its corresponding value.
The differences in the trends indicate that our method can adaptively learn compositionality levels for the phrases.
Table~\ref{tb:examples} shows the learned compositionality scores for the three groups of the examples along with the gold-standard scores given by the annotators.
The group (A) is considered to be consistent with the gold-standard scores, the group (B) is not, and the group (C) shows examples for which the difference between the compositionality scores of our results is large.

\paragraph{Characteristics of light verbs}
The verbs ``take'', ``make'', and ``have'' are known as {\it light verbs}
\footnote{In Section 5.2.2 in \newcite{light_verb}, the term {\it light verb} is used to refer to verbs which can be used in combination with some other element where their contribution to the meaning of the whole construction is reduced in some way.},
and the scoring function tends to assign low scores to light verbs.
In other words, our method can recognize that the light verbs are frequently used to form idiomatic (i.e. non-compositional) phrases.
To verify the assumption, we calculated the average compositionality score for each verb by averaging the compositionality scores paired with its candidate objects.
Here we used 135 verbs which take more than 30 types of objects in the BNC data.
Table~\ref{tb:ave_alpha} shows the 10 highest and lowest average scores with the corresponding verbs.
We see that relatively low scores are assigned to the light verbs as well as other verbs which often form idiomatic phrases.
As shown in the group (B) in Table~\ref{tb:examples}, however, light verb phrases are not always non-compositional.
Despite this, the learned function assigns low scores to compositional phrases formed by the light verbs.
These results suggest that using a more flexible scoring function may further strengthen our method.

\begin{table}[t]
  \begin{center}
{\small
    \begin{tabular}{lc|lc}
     \multicolumn{2}{l|}{Highest average scores} & \multicolumn{2}{l}{Lowest average scores}  \\ \hline \hline
	approve & 0.83 & bear & 0.37  \\
	reject & 0.72 & play & 0.38  \\
	discuss & 0.71 & have & 0.38  \\
	visit & 0.70 & make & 0.39  \\
	want & 0.70 & break & 0.40  \\
	describe & 0.70 & take & 0.40  \\
	involve & 0.69 & raise & 0.41  \\
	own & 0.68 & reach & 0.41  \\
	attend & 0.68 & gain & 0.42  \\
	reflect & 0.67 & draw & 0.42  \\ \hline
    \end{tabular}
}
    \caption{The 10 highest and lowest average compositionality scores with the corresponding verbs on the BNC data.}
    \label{tb:ave_alpha}
  \end{center}
\end{table}

\paragraph{Context dependence}
Both our method and the two datasets, VJ'05 and MC'07, assume that the compositionality score can be computed for each phrase with no contextual information.
However, in general, the compositionality level of a phrase depends on its contextual information.
For example, the meaning of the idiomatic phrase ``bear fruit'' can be compositionaly interpreted as ``to yield fruit'' for a plant or tree.
We manually inspected the BNC data to check whether the phrase ``bear fruit'' is used as the compositional meaning or the idiomatic meaning (``to yield results'').
As a result, we have found that most of the usage was its idiomatic meaning.
In the model training, our method is affected by the majority usage and fits the evaluation datasets where the phrase ``bear fruit'' is regarded as highly non-compositional.
Incorporating contextual information into the compositionality scoring function is a promising direction of future work.

\subsubsection{Effects of Ensemble}
We used the two different corpora for constructing the training data, and our method achieves the state-of-the-art results in all settings.
To inspect the results on VJ'05, we calculated the correlation score between the outputs from our results of the BNC and Wikipedia data.
The correlation score is 0.674 and that is, the two different corpora lead to reasonably consistent results, which indicates the robustness of our method.
However, the correlation score is still much lower than perfect correlation;
in other words, there are disagreements between the outputs learned with the corpora.
The group (C) in Table~\ref{tb:examples} shows such two examples.
In these cases, the ensemble technique is helpful in improving the results as shown in the examples.

Another interesting observation in our results is that the result of the ensemble technique outperforms that of the BNC-Wikipedia data as shown in Table~\ref{tb:compos}.
This shows that separately using the training corpora of different nature and then performing the ensemble technique can yield better results.
By contrast, many of the previous studies on embedding-based methods combine different corpora into a single dataset, or use multiple corpora just separately and compare them~\cite{hashimoto2015,muraoka2014,glove}.
It would be worth investigating whether the results in the previous work can be improved by ensemble techniques.

\section{Evaluation on the Phrase Embeddings}

\subsection{Evaluation Settings}

\paragraph{Dataset}
Next, we evaluated the learned embeddings on the transitive verb disambiguation dataset {\bf GS'11}\footnote{\url{http://www.cs.ox.ac.uk/activities/compdistmeaning/GS2011data.txt}} provided by \newcite{grefenstette2011}.
GS'11 consists of 200 pairs of transitive verbs and each verb pair takes the same subject and object.
For example, the transitive verb ``run'' is known as a polysemous word and this task requires one to identify the meanings of ``run'' and ``operate'' as similar to each other when taking ``people'' as their subject and ``company'' as their object.
In the same setting, however, the meanings of ``run'' and ``move'' are not similar to each other.
Each pair has multiple human ratings indicating how similar the phrases of the pair are.

\paragraph{Evaluation metric}
The evaluation was performed by calculating Spearman's rank correlation scores between the human ratings and the cosine similarity scores of $\bm{v}(SVO)$ in Eq.~(\ref{eq:svo_embed}).
Following the previous studies, we used the gold-standard ratings in two ways: averaging the human ratings for each SVO tuple (GS'11a) and treating each human rating separately (GS'11b).

\paragraph{Ensemble technique}
We used the same ensemble technique described in Section~\ref{subsec:compos_setting}.
In this task we produced two ensemble results: {\it Ensemble A} and {\it Ensemble B}.
The former used the averaged cosine similarity from the results of the BNC and Wikipedia data,
and the latter further incorporated the result of the BNC-Wikipedia data.

\paragraph{Baselines}
We compared our adaptive joint learning method with two baseline methods.
One is the method in \newcite{hashimoto2015} and it is equivalent to fixing $\alpha(VO)$ to $1$ in our method.
The other is fixing $\alpha(VO)$ to $0.5$ in our method, which serves as a baseline to evaluate how effective the proposed adaptive weighting method is.

\subsection{Results and Discussion}

\begin{table}[t]
  \begin{center}
{\small
    \begin{tabular}{l|c|c}
     Method & GS'11a & GS'11b \\ \hline \hline
     Proposed method (Wikipedia) & 0.598 & 0.461 \\
     Proposed method (BNC) & 0.595 & 0.463 \\
     Proposed method (BNC-Wikipedia) & 0.623 & 0.483 \\
	Proposed method (Ensemble A) & 0.661 & 0.511 \\
	Proposed method (Ensemble B) & 0.680 & 0.524 \\ \hline

	$\alpha(VO)=0.5$ (Wikipedia) & 0.491 & 0.386 \\
	$\alpha(VO)=0.5$ (BNC) & 0.599 & 0.462 \\
	$\alpha(VO)=0.5$ (BNC-Wikipedia) & 0.610 & 0.477 \\
	$\alpha(VO)=0.5$ (Ensemble A) & 0.612 & 0.474 \\
	$\alpha(VO)=0.5$ (Ensemble B) & 0.638 & 0.495 \\ \hline

	$\alpha(VO)=1$ (Wikipedia) & 0.576 & n/a \\
	$\alpha(VO)=1$ (BNC) & 0.574 & n/a \\ \hline
	\newcite{milajevs2014} & 0.456 & n/a \\
	\newcite{polajnar2014} & n/a & 0.370 \\
	\newcite{hashimoto2014} & 0.420 & 0.340 \\
	\newcite{polajnar2015} & n/a & 0.330 \\
	\newcite{grefenstette2011} & n/a & 0.210 \\ \hline
	Human agreement & 0.750 & 0.620 \\ \hline
    \end{tabular}
}
    \caption{Transitive verb disambiguation task.
The results for $\alpha(VO)=1$ are reported in \newcite{hashimoto2015}.}
    \label{tb:dis}
  \end{center}
\end{table}

\begin{table*}[t]
  \begin{center}
{\small
    \begin{tabular}{l|ll|l|l}
     & \multicolumn{2}{c}{Proposed method} & \multicolumn{1}{|c}{$\alpha(VO)=1$} & \multicolumn{1}{|c}{$\alpha(VO)=0.5$} \\ \hline \hline

	\multirow{5}{*}{take toll}
	& & put strain & deplete division & put strain \\
	& & place strain & necessitate monitoring & cause lack \\
	& $\alpha(\mathrm{take~toll})=0.11$ & cause strain & deplete pool & befall army \\
	& & have affect & create pollution & exacerbate weakness \\
	& & exacerbate injury & deplete field & cause strain \\ \hline

      \multirow{5}{*}{catch eye}
	& & catch attention & catch ear & grab attention \\
	& & grab attention & catch heart & make impression \\
	& $\alpha(\mathrm{catch~eye})=0.14$ & make impression & catch e-mail & catch attention \\
	& & lift spirit & catch imagination & become legend \\
	& & become favorite & catch attention & inspire playing \\ \hline

      \multirow{5}{*}{bear fruit}
	& & accentuate effect & bear herb & increase richness \\
	& & enhance beauty & bear grain & reduce biodiversity \\
	& $\alpha(\mathrm{bear~fruit})=0.19$ & enhance atmosphere & bear spore & fuel boom \\
	& & rejuvenate earth & bear variety & enhance atmosphere \\
	& & enhance habitat & bear seed & worsen violence \\ \hline

      \multirow{5}{*}{make noise}
	& & attack intruder & make sound & burn can \\
	& & attack trespasser & do beating & kill monster \\
	& $\alpha(\mathrm{make~noise})=0.33$ & avoid predator & get bounce & wash machine \\
	& & attack diver & get pulse & lightn flash \\
	& & attack pedestrian & lose bit & cook raman \\ \hline

      \multirow{5}{*}{buy car}
	& & buy bike & buy truck & buy bike \\
	& & buy machine & buy bike & buy instrument \\
	& $\alpha(\mathrm{buy~car})=0.71$ & buy motorcycle & buy automobile & buy chip \\
	& & buy automobile & buy motorcycle & buy scooter \\
	& & purchase coins & buy vehicle & buy motorcycle \\ \hline

    \end{tabular}
}
    \caption{Examples of the closest neighbors in the learned embedding space.
All of the results were obtained by using the Wikipedia data, and the values of $\alpha(VO)$ are the same as those in Table~\ref{tb:examples}.}
    \label{tb:ex_embed}
  \end{center}
\end{table*}

\subsubsection{Result Overview}

Table~\ref{tb:dis} shows our results and the state of the art, and our method outperforms almost all of the previous methods in both datasets.
Again, the ensemble technique further improves the results, and overall, Ensemble B yields the best results.

The scores in \newcite{hashimoto2015}, the baseline results with $\alpha(VO)=1$ in our method, have been the best to date.
As shown in Table~\ref{tb:dis}, our method outperforms the baseline results with $\alpha(VO)=0.5$ as well as those with $\alpha(VO)=1$.
We see that our method improves the baseline scores by adaptively combining compositional and non-compositional embeddings.
Along with the results in Table~\ref{tb:compos}, these results show that our method allows us to improve the composition function by jointly learning non-compositional embeddings and the scoring function for compositionality detection.

\subsubsection{Analysis of the Learned Embeddings}

We inspected the effects of adaptively weighting the compositional and non-compositional embeddings.
Table~\ref{tb:ex_embed} shows the five closest neighbor phrases in terms of the cosine similarity for the three idiomatic phrases ``take toll'', ``catch eye'', and ``bear fruit'' as well as the two non-idiomatic phrases ``make noise'' and ``buy car''.
The examples trained with the Wikipedia data are shown for our method and the two baselines, i.e., $\alpha(VO)=1$ and $\alpha(VO)=0.5$.
As shown in Table~\ref{tb:examples}, the compositionality levels of the first three phrases are low and their non-compositional embeddings are dominantly used to represent their meaning.

One observation with $\alpha(VO)=1$ is that head words (i.e. verbs) are emphasized in the shown examples except ``take toll'' and ``make noise''.
As with other embedding-based methods, the compositional embeddings are highly affected by their component words.
As a result, the phrases consisting of the same verb and the similar objects are often listed as the closest neighbors.
By contrast, our method flexibly allows us to adaptively omit the information about the component words.
Therefore, our method puts more weight on capturing the idiomatic aspects of the example phrases by adaptively using the non-compositional embeddings.

The results of $\alpha(VO)=0.5$ are similar to those with our proposed method, but we can see some differences.
For example, the phrase list for ``make noise'' of our proposed method captures offensive meanings, whereas that of $\alpha(VO)=0.5$ is somewhat ambiguous.
As another example, the phrase lists for ``buy car'' show that our method better captures the semantic similarity between the objects than $\alpha(VO)=0.5$.
This is achieved by adaptively assigning a relatively large compositionality score (0.71) to the phrase to use the information about the object ``car''.

We should note that ``make noise'' is highly compositional but our method outputs $\alpha(\mathrm{make~noise})=0.33$, and the phrase list of $\alpha(VO)=1$ is the most appropriate in this case.
Improving the compositionality detection function should thus further improve the learned embeddings.

\section{Related Work}

Learning embeddings of words and phrases has been widely studied,
and the phrase embeddings have proven effective in many language processing tasks, such as machine translation~\cite{cho2014,seq2seq}, sentiment analysis and semantic textual similarity~\cite{treeLSTM}.
Most of the phrase embeddings are constructed by word-level information via various kinds of composition functions like long short-term memory~\cite{LSTM} recurrent neural networks.
Such composition functions should be powerful enough to efficiently encode information about all the words into the phrase embeddings.
By simultaneously considering the compositionality of the phrases, our method would be helpful in saving the composition models from having to be powerful enough to perfectly encode the non-compositional phrases.
As a first step towards this purpose, in this paper we have shown the effectiveness of our method on the task of learning verb phrase embeddings.

Many studies have focused on detecting the compositionality of a variety of phrases~\cite{lin1999}, including the ones on verb phrases~\cite{diab2009,mccarthy2003} and compound nouns~\cite{farahmand2015,reddy2011}.
Compared to statistical feature-based methods~\cite{mccarthy2007,venkatapathy2005}, recent methods use word and phrase embeddings~\cite{kiela2013,yazdani2015}.
The embedding-based methods assume that word embeddings are given in advance and as a post-processing step, learn or simply employ composition functions to compute phrase embeddings.
In other words, there is no distinction between compositional and non-compositional phrases.
\newcite{yazdani2015} further proposed to incorporate latent annotations (binary labels) for the compositionality of the phrases.
However, binary judgments cannot consider numerical scores of the compositionality.
By contrast, our method adaptively weights the compositional and non-compositional embeddings using the compositionality scoring function.

\section{Conclusion and Future Work}
We have presented a method for adaptively learning compositional and non-compositional phrase embeddings by jointly detecting compositionality levels of phrases.
Our method achieves the state of the art on a compositionality detection task of verb-object pairs, and also improves upon the previous state-of-the-art method on a transitive verb disambiguation task.
In future work, we will apply our method to other kinds of phrases and tasks.

\section*{Acknowledgments}
We thank the anonymous reviewers for their helpful comments and suggestions.
This work was supported by CREST, JST.

\bibliography{bibtex.bib}

\begin{thebibliography}{}

\bibitem[\protect\citename{Bansal \bgroup et al.\egroup }2014]{bansal2014}
Mohit Bansal, Kevin Gimpel, and Karen Livescu.
\newblock 2014.
\newblock {Tailoring Continuous Word Representations for Dependency Parsing}.
\newblock In {\em Proceedings of the 52nd Annual Meeting of the Association for
  Computational Linguistics (Volume 2: Short Papers)}, pages 809--815.

\bibitem[\protect\citename{Cho \bgroup et al.\egroup }2014]{cho2014}
Kyunghyun Cho, Bart van Merrienboer, Caglar Gulcehre, Dzmitry Bahdanau, Fethi
  Bougares, Holger Schwenk, and Yoshua Bengio.
\newblock 2014.
\newblock {Learning Phrase Representations using RNN Encoder--Decoder for
  Statistical Machine Translation}.
\newblock In {\em Proceedings of the 2014 Conference on Empirical Methods in
  Natural Language Processing (EMNLP)}, pages 1724--1734.

\bibitem[\protect\citename{Church and Hanks}1990]{pmi}
Kenneth Church and Patrick Hanks.
\newblock 1990.
\newblock {Word Association Norms, Mutual Information and Lexicography}.
\newblock {\em Computational Linguistics}, 19(2):263--312.

\bibitem[\protect\citename{Diab and Bhutada}2009]{diab2009}
Mona Diab and Pravin Bhutada.
\newblock 2009.
\newblock {Verb Noun Construction MWE Token Classification}.
\newblock In {\em Proceedings of the Workshop on Multiword Expressions:
  Identification, Interpretation, Disambiguation and Applications}, pages
  17--22.

\bibitem[\protect\citename{Duchi \bgroup et al.\egroup }2011]{duchi2011}
John Duchi, Elad Hazan, and Yoram Singer.
\newblock 2011.
\newblock {Adaptive Subgradient Methods for Online Learning and Stochastic
  Optimization}.
\newblock {\em Journal of Machine Learning Research}, 12:2121--2159.

\bibitem[\protect\citename{Farahmand \bgroup et al.\egroup
  }2015]{farahmand2015}
Meghdad Farahmand, Aaron Smith, and Joakim Nivre.
\newblock 2015.
\newblock {A Multiword Expression Data Set: Annotating Non-Compositionality and
  Conventionalization for English Noun Compounds}.
\newblock In {\em Proceedings of the 11th Workshop on Multiword Expressions},
  pages 29--33.

\bibitem[\protect\citename{Grefenstette and Sadrzadeh}2011]{grefenstette2011}
Edward Grefenstette and Mehrnoosh Sadrzadeh.
\newblock 2011.
\newblock {Experimental Support for a Categorical Compositional Distributional
  Model of Meaning}.
\newblock In {\em Proceedings of the 2011 Conference on Empirical Methods in
  Natural Language Processing}, pages 1394--1404.

\bibitem[\protect\citename{Hashimoto and Tsuruoka}2015]{hashimoto2015}
Kazuma Hashimoto and Yoshimasa Tsuruoka.
\newblock 2015.
\newblock {Learning Embeddings for Transitive Verb Disambiguation by Implicit
  Tensor Factorization}.
\newblock In {\em Proceedings of the 3rd Workshop on Continuous Vector Space
  Models and their Compositionality}, pages 1--11.

\bibitem[\protect\citename{Hashimoto \bgroup et al.\egroup
  }2014]{hashimoto2014}
Kazuma Hashimoto, Pontus Stenetorp, Makoto Miwa, and Yoshimasa Tsuruoka.
\newblock 2014.
\newblock {Jointly Learning Word Representations and Composition Functions
  Using Predicate-Argument Structures}.
\newblock In {\em Proceedings of the 2014 Conference on Empirical Methods in
  Natural Language Processing (EMNLP)}, pages 1544--1555.

\bibitem[\protect\citename{Hochreiter and Schmidhuber}1997]{LSTM}
Sepp Hochreiter and J\"{u}rgen Schmidhuber.
\newblock 1997.
\newblock {Long Short-Term Memory}.
\newblock {\em Neural Computation}, 9(8):1735--1780.

\bibitem[\protect\citename{Kartsaklis \bgroup et al.\egroup }2012]{kart2012}
Dimitri Kartsaklis, Mehrnoosh Sadrzadeh, and Stephen Pulman.
\newblock 2012.
\newblock {A Unified Sentence Space for Categorical
  Distributional-Compositional Semantics: Theory and Experiments}.
\newblock In {\em Proceedings of the 24th International Conference on
  Computational Linguistics}, pages 549--558.

\bibitem[\protect\citename{Kartsaklis \bgroup et al.\egroup
  }2014]{kartsaklis2014}
Dimitri Kartsaklis, Nal Kalchbrenner, and Mehrnoosh Sadrzadeh.
\newblock 2014.
\newblock {Resolving Lexical Ambiguity in Tensor Regression Models of Meaning}.
\newblock In {\em Proceedings of the 52nd Annual Meeting of the Association for
  Computational Linguistics (Volume 2: Short Papers)}, pages 212--217.

\bibitem[\protect\citename{Kiela and Clark}2013]{kiela2013}
Douwe Kiela and Stephen Clark.
\newblock 2013.
\newblock {Detecting Compositionality of Multi-Word Expressions using Nearest
  Neighbours in Vector Space Models}.
\newblock In {\em Proceedings of the 2013 Conference on Empirical Methods in
  Natural Language Processing}, pages 1427--1432.

\bibitem[\protect\citename{Leech}1992]{bnc}
Geoffrey Leech.
\newblock 1992.
\newblock {100 Million Words of English: the British National Corpus}.
\newblock {\em Language Research}, 28(1):1--13.

\bibitem[\protect\citename{Levy and Goldberg}2014]{levy2014}
Omer Levy and Yoav Goldberg.
\newblock 2014.
\newblock {Dependency-Based Word Embeddings}.
\newblock In {\em Proceedings of the 52nd Annual Meeting of the Association for
  Computational Linguistics (Volume 2: Short Papers)}, pages 302--308.

\bibitem[\protect\citename{Lin}1999]{lin1999}
Dekang Lin.
\newblock 1999.
\newblock {Automatic Identification of Non-compositional Phrases}.
\newblock In {\em Proceedings of the 37th Annual Meeting of the Association for
  Computational Linguistics}, pages 317--324.

\bibitem[\protect\citename{McCarthy \bgroup et al.\egroup }2003]{mccarthy2003}
Diana McCarthy, Bill Keller, and John Carroll.
\newblock 2003.
\newblock {Detecting a Continuum of Compositionality in Phrasal Verbs}.
\newblock In {\em Proceedings of the ACL 2003 Workshop on Multiword
  Expressions: Analysis, Acquisition and Treatment}, pages 73--80.

\bibitem[\protect\citename{McCarthy \bgroup et al.\egroup }2007]{mccarthy2007}
Diana McCarthy, Sriram Venkatapathy, and Aravind Joshi.
\newblock 2007.
\newblock {Detecting Compositionality of Verb-Object Combinations using
  Selectional Preferences}.
\newblock In {\em Proceedings of the 2007 Joint Conference on Empirical Methods
  in Natural Language Processing and Computational Natural Language Learning},
  pages 369--379.

\bibitem[\protect\citename{Mikolov \bgroup et al.\egroup }2013]{word2vec}
Tomas Mikolov, Ilya Sutskever, Kai Chen, Greg~S Corrado, and Jeff Dean.
\newblock 2013.
\newblock {Distributed Representations of Words and Phrases and their
  Compositionality}.
\newblock In {\em Advances in Neural Information Processing Systems 26}, pages
  3111--3119.

\bibitem[\protect\citename{Milajevs \bgroup et al.\egroup }2014]{milajevs2014}
Dmitrijs Milajevs, Dimitri Kartsaklis, Mehrnoosh Sadrzadeh, and Matthew Purver.
\newblock 2014.
\newblock {Evaluating Neural Word Representations in Tensor-Based Compositional
  Settings}.
\newblock In {\em Proceedings of the 2014 Conference on Empirical Methods in
  Natural Language Processing}, pages 708--719.

\bibitem[\protect\citename{Mitchell and Lapata}2008]{lapata2008}
Jeff Mitchell and Mirella Lapata.
\newblock 2008.
\newblock {Vector-based Models of Semantic Composition}.
\newblock In {\em Proceedings of 46th Annual Meeting of the Association for
  Computational Linguistics: Human Language Technologies}, pages 236--244.

\bibitem[\protect\citename{Miyao and Tsujii}2008]{miyao2008}
Yusuke Miyao and Jun'ichi Tsujii.
\newblock 2008.
\newblock {Feature Forest Models for Probabilistic {HPSG} Parsing}.
\newblock {\em Computational Linguistics}, 34(1):35--80, March.

\bibitem[\protect\citename{Muraoka \bgroup et al.\egroup }2014]{muraoka2014}
Masayasu Muraoka, Sonse Shimaoka, Kazeto Yamamoto, Yotaro Watanabe, Naoaki
  Okazaki, and Kentaro Inui.
\newblock 2014.
\newblock {Finding The Best Model Among Representative Compositional Models}.
\newblock In {\em Proceedings of the 28th Pacific Asia Conference on Language,
  Information, and Computation}, pages 65--74.

\bibitem[\protect\citename{Newton}2006]{light_verb}
Mark Newton.
\newblock 2006.
\newblock {\em {Basic English Syntax with Exercises}}.
\newblock B\"{o}lcs\'{e}sz Konzorcium.

\bibitem[\protect\citename{Noreen}1989]{bootstrap}
Eric~W. Noreen.
\newblock 1989.
\newblock {\em {Computer-Intensive Methods for Testing Hypotheses: An
  Introduction}}.
\newblock Wiley-Interscience.

\bibitem[\protect\citename{Pennington \bgroup et al.\egroup }2014]{glove}
Jeffrey Pennington, Richard Socher, and Christopher Manning.
\newblock 2014.
\newblock {Glove: Global Vectors for Word Representation}.
\newblock In {\em Proceedings of the 2014 Conference on Empirical Methods in
  Natural Language Processing (EMNLP)}, pages 1532--1543.

\bibitem[\protect\citename{Pham \bgroup et al.\egroup }2015]{cphrase}
Nghia~The Pham, Germ\'{a}n Kruszewski, Angeliki Lazaridou, and Marco Baroni.
\newblock 2015.
\newblock {Jointly optimizing word representations for lexical and sentential
  tasks with the C-PHRASE model}.
\newblock In {\em Proceedings of the 53rd Annual Meeting of the Association for
  Computational Linguistics and the 7th International Joint Conference on
  Natural Language Processing (Volume 1: Long Papers)}, pages 971--981.

\bibitem[\protect\citename{Polajnar \bgroup et al.\egroup }2014]{polajnar2014}
Tamara Polajnar, Laura Rimell, and Stephen Clark.
\newblock 2014.
\newblock {Using Sentence Plausibility to Learn the Semantics of Transitive
  Verbs}.
\newblock In {\em Proceedings of Workshop on Learning Semantics at the 2014
  Conference on Neural Information Processing Systems}.

\bibitem[\protect\citename{Polajnar \bgroup et al.\egroup }2015]{polajnar2015}
Tamara Polajnar, Laura Rimell, and Stephen Clark.
\newblock 2015.
\newblock {An Exploration of Discourse-Based Sentence Spaces for Compositional
  Distributional Semantics}.
\newblock In {\em Proceedings of the First Workshop on Linking Computational
  Models of Lexical, Sentential and Discourse-level Semantics}, pages 1--11.

\bibitem[\protect\citename{Reddy \bgroup et al.\egroup }2011]{reddy2011}
Siva Reddy, Diana McCarthy, and Suresh Manandhar.
\newblock 2011.
\newblock {An Empirical Study on Compositionality in Compound Nouns}.
\newblock In {\em Proceedings of 5th International Joint Conference on Natural
  Language Processing}, pages 210--218.

\bibitem[\protect\citename{Socher \bgroup et al.\egroup }2012]{socher2012}
Richard Socher, Brody Huval, Christopher~D. Manning, and Andrew~Y. Ng.
\newblock 2012.
\newblock {Semantic Compositionality through Recursive Matrix-Vector Spaces}.
\newblock In {\em Proceedings of the 2012 Joint Conference on Empirical Methods
  in Natural Language Processing and Computational Natural Language Learning},
  pages 1201--1211.

\bibitem[\protect\citename{Sutskever \bgroup et al.\egroup }2014]{seq2seq}
Ilya Sutskever, Oriol Vinyals, and Quoc~V Le.
\newblock 2014.
\newblock {Sequence to Sequence Learning with Neural Networks}.
\newblock In {\em Advances in Neural Information Processing Systems 27}, pages
  3104--3112.

\bibitem[\protect\citename{Tai \bgroup et al.\egroup }2015]{treeLSTM}
Kai~Sheng Tai, Richard Socher, and Christopher~D. Manning.
\newblock 2015.
\newblock {Improved Semantic Representations From Tree-Structured Long
  Short-Term Memory Networks}.
\newblock In {\em Proceedings of the 53rd Annual Meeting of the Association for
  Computational Linguistics and the 7th International Joint Conference on
  Natural Language Processing (Volume 1: Long Papers)}, pages 1556--1566.

\bibitem[\protect\citename{Van~de Cruys}2014]{vande2014}
Tim Van~de Cruys.
\newblock 2014.
\newblock {A Neural Network Approach to Selectional Preference Acquisition}.
\newblock In {\em Proceedings of the 2014 Conference on Empirical Methods in
  Natural Language Processing (EMNLP)}, pages 26--35.

\bibitem[\protect\citename{Venkatapathy and Joshi}2005]{venkatapathy2005}
Sriram Venkatapathy and Aravind Joshi.
\newblock 2005.
\newblock {Measuring the Relative Compositionality of Verb-Noun (V-N)
  Collocations by Integrating Features}.
\newblock In {\em Proceedings of Human Language Technology Conference and
  Conference on Empirical Methods in Natural Language Processing}, pages
  899--906.

\bibitem[\protect\citename{Yazdani \bgroup et al.\egroup }2015]{yazdani2015}
Majid Yazdani, Meghdad Farahmand, and James Henderson.
\newblock 2015.
\newblock {Learning Semantic Composition to Detect Non-compositionality of
  Multiword Expressions}.
\newblock In {\em Proceedings of the 2015 Conference on Empirical Methods in
  Natural Language Processing}, pages 1733--1742.

\end{thebibliography}
\bibliographystyle{acl2016}

\end{document}